\documentclass{article}
\usepackage[numbers,sort&compress]{natbib}
\usepackage[final]{neurips_2024}

\usepackage[utf8]{inputenc} % allow utf-8 input
\usepackage[T1]{fontenc}    % use 8-bit T1 fonts
\usepackage{hyperref}       % hyperlinks
\usepackage{url}            % simple URL typesetting
\usepackage{booktabs}       % professional-quality tables
\usepackage{amsfonts}       % blackboard math symbols
\usepackage{nicefrac}       % compact symbols for 1/2, etc.
\usepackage{microtype}      % microtypography
\usepackage[table]{xcolor}
\usepackage{arydshln}

\usepackage{graphicx}
\usepackage{algorithm}
\usepackage{algpseudocode}
\usepackage{multirow}
\usepackage{makecell}
\usepackage{amsmath}
\usepackage{amssymb}% http://ctan.org/pkg/amssymb
\usepackage{pifont}% http://ctan.org/pkg/pifont

\usepackage{subfigure}
\usepackage{caption}
\usepackage{subcaption}
\usepackage{graphicx,wrapfig}
\usepackage{enumitem,kantlipsum}
\hypersetup{colorlinks, citecolor=blue}

\title{Identifying Differential Patient Care \\Through Inverse Intent Inference}

\author{%
  Hyewon Jeong
  \thanks{Correspondence to: hyewonj@mit.edu} 
    \\
  MIT\\
  \texttt{hyewonj@mit.edu}\\
   \And
   Siddharth Nayak \\
   MIT\\
   \texttt{sidnayak@mit.edu} \\
   \And
   Taylor Killian \\
   MIT, University of Toronto\\
   \texttt{twk@mit.edu} \\
   \And
   Sanjat Kanjilal \\
   Harvard Medical School\\
   \texttt{skanjilal@bwh.harvard.edu}\\
}

\begin{document}

\maketitle
\begin{abstract}
Sepsis is a life-threatening condition defined by end-organ dysfunction due to a dysregulated host response to infection. Although the Surviving Sepsis Campaign has launched and has been releasing sepsis treatment guidelines to unify and normalize the care for sepsis patients, it has been reported in numerous studies that disparities in care exist across the trajectory of patient stay in the emergency department and intensive care unit. Here, we apply a number of reinforcement learning techniques including behavioral cloning, imitation learning, and inverse reinforcement learning, to learn the optimal policy in the management of septic patient subgroups using expert demonstrations. Then we estimate the counterfactual optimal policies by applying the model to another subset of unseen medical populations and identify the difference in cure by comparing it to the real policy. Our data comes from the sepsis cohort of MIMIC-IV and the clinical data warehouses of the Mass General Brigham healthcare system. The ultimate objective of this work is to use the optimal learned policy function to estimate the counterfactual treatment policy and identify deviations across sub-populations of interest. We hope this approach will help us identify any disparities in care and also changes in cure in response to the publication of national sepsis treatment guidelines.
\end{abstract}

\section{Introduction}

Sepsis is a severe reaction by the human body to infection and is associated with significant morbidity and mortality \cite{singer2016third}. National practice guidelines in the United States for the management of sepsis were updated in 2016 to normalize care patterns \cite{singer2016third, rhodes2017surviving}, and the Surviving Sepsis Campaign (SSC) has released the updates in international guidelines \cite{evans2021surviving}. Societal factors of the patients should not affect the care and outcome of patients if the treatment has been normalized following the guidelines, however, there remain notable disparities in outcomes by gender \cite{sunden2020sex} and race/ethnicity \cite{dimeglio2018factors, jones2017racial, moore2015black}. It has shown that male patients on average require more resources with higher nurse workloads per admission \cite{samuelsson2015gender} and the female patients with higher severity of illness are less likely to receive advanced life support measures (e.g., mechanical ventilation) than the male cohort \cite{fowler2007sex}. The bias in care has also been reported in treating other disease as well \cite{regitz2006therapeutic, regitz2010heart}. This might be in part due to implicit bias of healthcare professionals in making treatment decisions against different racial and gender groups \cite{fitzgerald2017implicit}.

\begin{figure*}[ht!]
	\centering
	\includegraphics[width=\textwidth]{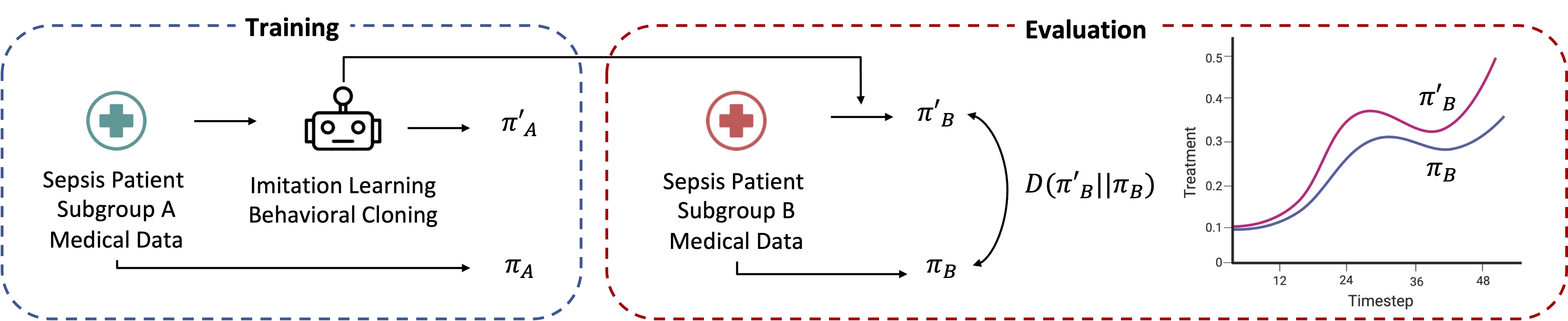}
    \caption{Analytic approach to identify differences in sepsis treatment policies across different patient subgroups with Imitation Learning and Behavioral Cloning. We let model agent learn the expert trajectory in one patient subgroup, then apply this learned agent to another patient subgroup to get the discrepancy between counterfactual policy $\pi'_B = \pi'(\textbf{z}_i|A=B)$ where $\textbf{z}_i$ is the set of state transition, conditioned on the patient attribute $A=B$, and original policy $\pi_B$.}
	\label{concept}
\end{figure*}

Sepsis has been the focus of intense research in the field of machine learning with the primary aim being the ability to predict the onset of disease and to identify the optimal treatment policies for this complex condition. Advances in the scale and granularity of electronic health record data offer the opportunity to apply reinforcement learning (RL) to understand clinician diagnostic and treatment policies for this complex condition \cite{RL_sepsis_Raghu17, kernel_RL_sepsis_Peng17, RL_ICU_PrasadCCDE17}, which can be used to understand the factors that drive disparities in sepsis care. The fundamental problem in using RL to model sepsis is that the reward function is unknown and involves tradeoffs between competing outcomes. In this work, we utilize a set of reinforcement learning methods, on- and off-policy Imitation Learning (IL) \cite{ng2000algorithms, abbeel2004apprenticeship, ziebart2008maximum, ho2016generative, peng2018variational} to learn and map state-action pairs from retrospective data, thereby learning the expert policy (Figure \ref{concept}, Training). 

With the learned models on the subset of the dataset (e.g., patient cure data under guideline A), we can estimate the treatment policy with this learned model on another subgroup (e.g., medical data under guideline B) (Figure \ref{concept}, Evaluation). This is the estimated \textit{counterfactual} policy a patient is receiving if the patient was under guideline A given the same feature input. We analyze counterfactual fairness of the estimated policy given different demographic aspects of patients (gender, race), national treatment guidelines \cite{singer2016third}, and the insurance plan a patient is holding. This helps us identify whether treatment strategies employed by healthcare providers differ by subgroups as identified from the simple plotting of the treatment trajectory itself (Appendix Figure \ref{action_trajectory}) and previous works \cite{sunden2020sex, samuelsson2015gender, fowler2007sex}.

We apply this approach to two large and independent datasets to counterfactually estimate the differences in care. Two datasets include the sepsis patient cohort with the clinical features, where the first dataset is from Medical Information Mart for Intensive Care (MIMIC)-IV \cite{johnson2020mimic} and the second dataset is from the clinical data warehouse of the Mass General Brigham (MGB) healthcare system. MIMIC-IV cohort includes patient clinical features related to sepsis and hourly treatment around the time window where the patient has been diagnosed with sepsis while the patient is admitted to the Intensive Care Unit (ICU). MGB cohort has records of patients with community-onset adult sepsis event (ASE), including the patient trajectory from the time of arrival of the patient in the emergency room until hospital discharge across 12 hospitals in the New England area from 2015 through 2022. 

Our approach will help identify the existence of differential treatment policies across subgroups of patients with sepsis, where in most cases the counterfactual situation cannot be simulated or collected. By pinpointing disparities in care across gender and ethnic groups, particularly in the pre-ICU triage, we can inform and target interventions to promote health equity. Furthermore, our approach can also identify the shifts in treatment patterns associated with changes in national guidelines, which can help estimate their true impact on clinical care. Interrogation of individual trajectories may quantify the distributional deviation from the expert policy to the estimated counterfactual optimal strategies, which can be captured as a notion of surprise \cite{achiam2017surprise} or distributional deviation \cite{zhao2021comparing}. We believe identifying different subsets of policies within a patient cohort can further aid in the interpretation of feature sets that are the primary drivers for change in the action trajectory.

\section{Related Works}
\subsection{Imitation Learning}
Imitation learning (IL) \cite{ross2011reduction, ross2014reinforcement} aims to mimic expert behavior in given tasks, which has been demonstrated with various sequential decision making tasks including autonomous driving \cite{pomerleau1988alvinn}, helicopter acrobatics \cite{coates2008learning}, ghosting \cite{le2017data}, speech animation \cite{taylor2017deep}. Imitation learning facilitates teaching machine learning agents complex tasks with minimal expert knowledge of the tasks and has been used for inferring human intent \cite{ziebart2009planning}. IL lets a machine learning model or an agent trained to perform tasks from demonstrations by learning a mapping between observations and actions. Broadly, there are off- and on-policy categories in imitation learning: Behavioral cloning (off-policy IL) and Inverse Reinforcement Learning (on-policy IL).

Behavioral cloning (BC) \cite{bc1, bc2} is an off-policy imitation learning, where the agent receives states and actions from the expert dataset and then a regressor or a classifier is learned to replicate the policy used in the expert dataset. A major advantage of this method is the capability of imitating the expert demonstrations without requiring additional interactions with the environment. BC has been used to train complex control policies acquired from human demonstrations to pilot quadrotor drones \cite{bc_quadrotor} as well as the autonomous operation of passenger vehicles \cite{bc_autodrive}. 

As one of the on-policy IL, Inverse reinforcement learning (IRL) \cite{ng2000algorithms, abbeel2004apprenticeship, ziebart2008maximum} techniques seek to learn a reward function that has the maximum value for the demonstrated actions. The learned reward function is then used in combination with standard reinforcement learning methods to find an imitation policy. IRL has even been scaled to learn reward functions under unknown dynamics in high-dimensional continuous control settings \cite{finn_irl_robotics}. 

Building on top of the success of Generative Adversarial Networks (GANs) \cite{gans}, Generative Adversarial Imitation Learning (GAIL) \cite{ho2016generative} trains generators such that they have similar behaviors to the given experts, and discriminators judge whether the behaviors look like the experts with a defined reward function. Variational Adversarial Imitation Learning (VAIL) \cite{peng2018variational} was proposed to tackle training stability issues of adversarial models, enforcing a constraint on the mutual information between observations and the discriminator's internal representation. This helps modulate the discriminator's accuracy and maintain useful and informative gradients.

\subsection{Reinforcement Learning for Clinical Decision Making}
Reinforcement learning has been used to find optimal treatments in the Intensive Care Unit (ICU) \cite{RL_sepsis_Raghu17}, where the policies were modeled as continuous state space with clinically-guided reward function. They perform experiments with Double Q-Learning \cite{double_q_learning} and Dueling Double Q-Networks \cite{wang2016dueling}. RL has also been used to identify treatments to avoid high-risk treatment among patients with sepsis \cite{fatemi2021medical}. In another work \cite{kernel_RL_sepsis_Peng17}, a mixture-of-experts framework has been adopted to personalize sepsis treatment where it alternates between a neighbor-based (kernel) model and a deep reinforcement learning based model depending on the patient's history. Similarly, off-policy reinforcement learning algorithms have also been applied \cite{RL_ICU_PrasadCCDE17} to determine the best action at a given patient state from sub-optimal historical ICU data, which helps improved outcomes in terms of minimizing reintubation rates and regulation of physiological stability. A recent work adopts adversarial IRL to enforce learning reward structure in a more robust way such that it can be used to better understand vasopressor and IV fluid administration in ICU \cite{srinivasan2020interpretable}. Building on top of these previous works of RL applied to clinical decision making tasks, we would like to further evaluate and discover the subgoals or intent of experts by counterfactual estimation of policy.

\section{Approach}
\subsection{Learning the optimal treatment selection}
\label{learn}
We have a set of data with patient trajectory $D = \{(s_i, a_i, s_{i+1})\}^N_{i=1}$ compiled for clinical risk prediction task $\mathcal{T}$ where $s_i$ is clinical status of $i$-th patient, and $a_i$ is the course of action (treatment) on patient $i$. Here, each state (patient instance) $s_i = [\mathbf{w}_1, \mathbf{w}_2, \dots, \mathbf{w}_M]$ contains the time-series input of $M$ clinical features each $\mathbf{w}_{j \in [0,M]}$ sampled for $T$ steps around the time a patient has diagnosed with sepsis \cite{killian2020empirical}. Then we utilize a set of methods including off- and on-policy IL, Behavioral Cloning, and Inverse Reinforcement Learning (IRL). 

We formulate our framework as a Markov Decision Process (MDP) defined with a sequence of transitions $\textbf{z}_i = \{(s_t, a_t, r_{t+1}, s_{t+1})\}_{t \geq 0}$ per each patient $\textbf{z}_i$, where the reward $r_{t+1}$ at each timestep can be obtained by the learned reward function from IL framework. With the IL framework, we learn a function $\pi(\textbf{z}_i)$ which fits the expert trajectory of patient subgroup $c_k$, $D_{c_k} = \{(\textbf{z}_i, a_i)\}^N$, where $\textbf{z}_i = \{(s_t, a_t, s_{t+1})\}_{t \geq 0}$. Thus the learned policy can be written using the fact that it has been obtained by adjusting for the patient attribute $A=k$: $\pi(\textbf{z}_i|A=k) = \pi(\textbf{z}^{c_k}_i)$. Note that our framework is based on observational data in a partially-observed MDP (POMDP) setting which fits well with medical time-series data. 

Then we evaluate this learned framework on the record of another subgroup of patient $c_l$, $D_{c_l} = \{(\textbf{z}^{c_l}_i, a^{c_l}_i)\}^N$. After obtaining the mapped counterfactual policy $\pi^*(\textbf{z}^{c_l}_i) = \pi^*(\textbf{z}_i|A=l)$ adjusted for the patient attribute $A=a$, given state $s_t$ for each inter-sectional sub-groupings $c$, we then assess variations among them (Section \ref{trajectory}).

\textbf{Behavioral Cloning} As our baseline model, behavioral cloning learns a policy where the actions are as close as possible to the expert actions $a$. To optimize the model, the loss function used to learn from the dataset depends on the type of output. For continuous action spaces, we use the Root Mean Square Error (RMSE) for a batch size of n: 
\begin{align*}
    \mathcal{L} = \sqrt{\frac{\sum\limits_{i=1}^{n}\vert\vert a - \pi(\textbf{z}|A=k)\vert\vert^2}{n}}
\end{align*}
For discrete action spaces, we use the negative log-likelihood loss function: 
\begin{align*}
\mathcal{L}= \sum\limits_{i=1}^{n}\log{\sigma(\pi(\textbf{z}_i|A=k))_{a^*_i}},
\end{align*}
where $\sigma$ is the softmax function which ensures that the output of the neural network for the multi-class prediction problem is a probability distribution. With the learned behavioral cloning model, we obtain the counterfactual policy $\pi^*(\textbf{z}^{c_l}_i) = \pi^*(\textbf{z}_i|A=l)$ on a new subset of patient group corresponding to the patient attribute $A=l$ for further evaluation.

\textbf{State-Transition Environment Model} To evaluate the learned policies using offline-RL methods, we create an environment with the above MDP $\textbf{z}_i = \{(s_t, a_t, r_{t+1}, s_{t+1})\}_{t \geq 0}$ to fit the state-transition function $f$ \cite{jiang2016MBEnv, raghu2018model}. Note that we use the learned IRL model to get the reward values and model this environment function $f(h_t; \theta)$ only for fitting a transition function, with the setting adopted from Raghu et al., \cite{raghu2018model}. The state-transition environment model $f(h_t ; \theta)$ gets the input state-action pairs $h_t = g(\{s_t, s_{t-1}, s_{t-2}$, $a_t, a_{t-1}, a_{t-2}\})$ where $g$ concatenates the state action pairs of previous three timesteps and $f$ outputs the changes in the physiological states of patients, $\Delta_t=s_{t+1}-s_t$. $\Delta$ is then added to the current state $s_t$ to get the next stage which is used to construct state transition environment $\textbf{z}_i = \{(s_t, a_t, r_{t+1}, s_{t+1})\}_{t \geq 0}$.

\textbf{Inverse Reinforcement Learning}
We use IRL in combination with the environment modeling described above to learn the reward function given state action pairs $\{(s_i, a_i, s_{i+1})\}^N_{i=1}$. The reward function is defined specifically to our task of interest, using the outcome of the patients (Section \ref{data} for detail). The IRL module is formulated as learning a reward function $R^*(s)$ such that $\mathbb{E}[\sum_{t=0}^{\infty}\gamma^t R^*(s_t)|\pi_E] \geq \mathbb{E}[\sum_{t=0}^{\infty}\gamma^t R^*(s_t)|\pi], \forall{\pi}$ where $\pi_E$ is the expert policy.
After we obtain the reward function from IRL, we propose to use offline RL \cite{fujimoto2021BCQ, kumar2020CQL} which can is used to learn $a_t \sim \pi(a|s_t, A=k)$ where we obtain mapped action $a_t$ given state $s_t$ and inter-sectional sub-groupings with patient attribute $A=k$. We get the mapped counterfactual action $a_{t, l}^* = \pi^*(a|s_t, A=l)$, which is the output of the policy model $\pi$, from a new subgroup $c_l$ to assess variations across them.

\textbf{Generative Adversarial Imitation Learning (GAIL)}
\cite{ho2016generative} We use GAIL to fit a policy over the expert dataset where we improve both the discriminator and policy simultaneously over the iteration. The discriminator $D$ parameters are updated to differentiate between the expert trajectories $\tau_E \sim \pi_E$ with $\tau_E = (s_1, a_1, s_2, a_2, \cdots)$ and generated pairs $\tau_i \sim \pi_{\theta_i}$ with the loss below.
\begin{align*}
    \hat{\mathbb{E}}_{\pi}[\nabla_w \log(D_w(s, a))]+\hat{\mathbb{E}}_{\pi_E}[\nabla_w \log(1-D_w(s, a))]
\end{align*}
Then this discriminator loss is used to update the generated policy using the Trust Region Policy Optimization (TRPO) rule where $Q(\bar{s}, \bar{a})=\hat{\mathbb{E}}_{\pi_\theta}[\log (D_w(s, a)) \mid s_0=\bar{s}, a_0=\bar{a}]$ and causal entropy $H(\pi_\theta)$ be the policy regularizer.
\begin{align*}
    \hat{\mathbb{E}}_\pi[\nabla_\theta \log \pi_\theta(a \mid s) Q(s, a)]-\lambda \nabla_\theta H(\pi_\theta)
\end{align*}
The discriminator also serves as the reward function for policy improvement by judging whether the behaviors look like the expert data. We get the counterfactual policy in a new patient attribute group in the same way as explained in the IRL section above.

\subsection{Identifying the deviation to optimal behavior trajectory}
\label{trajectory}

Given a model learned the expert policy $\pi_E(\textbf{z}_i)$ from a subset of patient $D_{c_k}$, we can identify the deviation of action trajectory in another subset of patient $D_{c_l}$. This deviation from using discrepancy measures: Integral Probability Measures (IPM) \cite{muller1997integral} such as Wasserstein distance, Maximum mean discrepancy (MMD) \cite{rao1982diversity, rao1987differential, gretton2012kernel, sinn2012detecting}, and \textit{f}-divergences such as Kullback–Leibler divergence (KL divergence) or Jensen Shannon divergence. In the case of comparing decision trajectories \cite{zhao2021comparing}, we can define discrepancy term using KL divergence as: $D_{KL}(\mathbf{a_{c_l}}||\mathbf{\pi^*}) \approx \mathbb{E}_{z_l}[\pi_k^*(z_l)||\pi_E(z_l)]$ where we can rewrite the LHS of the equation as: $D_{KL}(\mathbf{a_{c_l}}||\mathbf{\pi^*}) = D_{KL}(\mathbf{a_{c_l}}||\pi^*(\textbf{z}^{c_l}_i)) = D_{KL}(\mathbf{a_{c_l}}||\pi^*(\textbf{z}_i|A=l))$
The evaluated discrepancy can be further used to identify the features generating this divergence, fair evaluation of subset, or can be used as a surprisal incentive to explore \cite{achiam2017surprise}. 

\section{Experiments}

\subsection{Datasets}
\label{data}

\textbf{MIMIC-IV Sepsis Cohort} \\
We develop experiments using a dataset derived from out of the MIMIC-IV \cite{johnson2020mimic, goldberger2000physiobank} database, which contains electronic health records (EHR) from the intensive care unit from the Beth Israel Deaconess Medical Center (BIDMC). The sepsis cohort in a MIMIC-IV dataset is defined with the inclusion criteria for patients meeting the sepsis-3 definition where the patients were diagnosed with any type of infection and further have more than two sepsis-related organ failures (SOFA). We summarize common features predictive of patient septic status in Table \ref{feature}. We removed patients who stayed less than 12 hours in ICU and also removed patients who did not have recorded vital signs for more than 6 hours. We only included patients over the age $18$ who were initially admitted to the Medical Intensive Care Unit (MICU) for the homogeneity of our patient cohort. Patients from other ICUs might develop sepsis from inherently different causes (e.g. may arise after surgical intervention). This filtering choice ensures some measure of homogeneity in the possible causes and observed treatment strategies for sepsis. We include the first ICU visit per each patient, and hourly sample patient records, up to 24 hours before and 48 hours after presumed sepsis onset. Using the exclusion criteria described above, we compiled an experimental cohort of $9,306$ unique patient encounters, spanning a total of 72 hours centered on a presumed onset of sepsis. 

We select 40 sepsis-related features and 2 patient attributes (Table \ref{feature}), then hourly sampled them to construct the dataset. Every non-binary feature and action was z-normalized while preprocessing, and some of the features with skewed distribution were log-normalized. We filled the Not Available (N/A) dataset in between observations with forward-filling (carry-forward) imputation and linear interpolation method. Any observations before the first data point were filled with the average feature value. The dataset was randomly split to approximately 6:2:2 for training/validation/testing. We then define both continuous actions and binned categorical actions following previous literature \cite{johnson2018mimic, gottesman2020interpretable}. Continuous action space is used for behavioral cloning regression and discrete categorical action space is used for classification.

\begin{table*}[ht!]
	\small
	\centering
	\caption{Selected features to construct the sepsis cohort dataset. Total 42 features including 2 patient attributes were selected for the dataset, which corresponds to each of the gross category summarized as below.}
	\begin{tabular}{cc}
		\toprule
		\textbf{Feature Category} & \textbf{Features}\\
		\midrule
        \textbf{Patient Attributes} & gender, ethnicity \\\midrule
        Demographic Features & age\\\midrule
        General Features & height, weight\\\midrule
        \multirow{2}{*}{Vital Signs} & heart rate, systolic blood pressure, diastolic blood pressure,\\
        & mean blood pressure, respiratory rate, temperature\\\midrule
        Blood Gas & pH, base excess\\ \midrule
        Complete Blood Count (CBC) & Hematocrit, Hemoglobin, Platelet, White blood Cell count (WBC) \\ \midrule
        Blood Chemistry & Chloride, Calcium, Potassium, Sodium, Lactate \\ \midrule
        Blood coagulation & PT, aPTT, INR \\ \midrule
        Respiratory System & Sa$O_2$, Sp$O_2$, Pa$O_2$, PaC$O_2$, Fi$O_2$, Pa$O_2$/Fi$O_2$ ratio\\\midrule
        Renal System & Urine Nitrogen (BUN), Creatinine, Albumin, Aniongap, Bicarbonate\\ \midrule
        Liver Function Test & Bilirubin, ALT, AST\\ \midrule
        Input/Output & Urine Output \\ \midrule
        Mental Status & Glasgow Coma Scale (GCS) \\
		\bottomrule
	\end{tabular}
\label{feature}
\end{table*}

\textbf{Massachusetts General Brigham Sepsis Cohort} \\
Data is provided, by agreement, from the clinical data warehouse of Mass General Brigham (MGB) where a total of $54,594$ patients were identified with community-onset adult sepsis event (ASE) from $267,790$ patients who meet the criteria set by the Centers for Disease Control and Prevention (CDC). MGB is the largest healthcare network in the New England area, encompassing 12 hospitals serving a highly diverse population, with data from 2015 through the present. Patient records were sampled hourly for 72 hours after the patient entered the emergency room or ICU. The same features are used as derived for the MIMIC-IV dataset, where creatinine has replaced with calculated Estimated Glomerular Filtration Rate (eGFR) value to better record the renal function of a patient. In addition to the two patient attributes gender and ethnicity, the admission year (to adjust for the guideline the patient was treated under) and the insurance plan a patient is holding have collected.

\textbf{Continuous Action Space} \\
\label{conti_action}
We define the continuous action on septic patients to be the amount of IV fluid bolus (Normal Saline, Dextrose 5\% in Water (D5W)) and vasopressors (Norepinephrine, Phenylephrine, Vasopressin, Dobutamine, Milrinone) administered to a patient at the specific time point. Phenylephrine, Dopamine, and Vasopressin doses were converted to be equivalent to standard Norepinephrine amounts following the MIMIC code repository \cite{johnson2018mimic}. To calculate the equivalent dose for vasopressors \cite{goradia2021vasopressor}, we multiplied the dose of each medication by a specific value, where the phenylephrine doses were divided by 10, Dopamine doses were divided by 100, and Vasopressin doses were multiplied by 2.5. 
 
\textbf{Discrete Action Space} \\
\label{discrete_action}
We then bin the combined administered doses of each treatment (IV fluid bolus and Vasopressors) into 5 categories where the cutoff is defined by the quantiles of nonzero continuous medication (0\%, 25\%, 50\%, 75\%) so that the full action space consists of 25 discrete actions (5 bins for fluid in combination with 5 bins for vasopressors), following Komorowski et al., \cite{komorowski2018artificial}.

\textbf{Reward model}\\
We used the outcome (mortality, hypertensive crisis, hypotensive crisis) of sepsis patients to be the proxy of reward modeling. For the mortality proxy flag, we sparsely modeled the reward function by penalizing the mortality at the time of the event in the intermediate timestep (1-71 timestep, zero reward when the patient is alive), and for the final timestep, the agent is penalized for the mortality and rewarded when a patient is alive. The reward for hyper/hypotensive crisis is defined using mean arterial pressure (MAP) and systolic blood pressure (SBP) where for the timestep with MAP < 60mmHg and SBP > 180mmHg, the model is penalized for hypotensive crisis and hypertensive crisis, respectively. For the normal MAP range of 60-80 mmHg, the model gets the positive reward.

\subsection{Details on Implementation and Experimental Setting}
\label{implementation}
\textbf{State-Transition Model}\\
We used Long-Short Term Memory \cite{hochreiter1997long} network for the architecture of the state-transition model, where the input state-action pairs of the previous three steps predict the state at the next timestep.

\textbf{Behavioral Cloning}\\
All experiments were run for 300 epochs where dataloader feeds the batch of size 64 to model (linear or CNN classifier/regressor) per iteration. Baseline architectures, such as the Feed-Forward Model (FFW), two-layer Convolutional Neural Network (CNN), and ResNet18 \cite{he2016deep} have been adopted for behavioral cloning and environment modeling. FFW network is a linear classifier with three linear layers, batch normalization, and rectified linear unit (ReLU) was used for the activation function. Flattened input, Batch $\times$ (timestep $\times$ features) fed into the model to output Batch $\times$ timestep continuous or discrete actions. For CNN and ResNet models, 3D input Batch $\times$ timestep $\times$ features fed into the model to output Batch $\times$ timestep continuous or discrete actions. Several hyperparameters (numbers of layers, hidden dimensions) were tuned for the best performance.

\textbf{GAIL}\\
The GAIL \cite{ho2016generative} model \footnote{Code from \href{https://github.com/uidilr/deepirl_chainer}{https://github.com/uidilr/deepirl\_chainer}} includes a policy network and a discriminator network. The policy network is a 2-layer feed-forward network with 200 neurons per layer. The discriminator network is a 3-layer feed-forward network with 64 neurons per layer. We use a learning rate of $3e^{-4}$ and a batch size of 64.

\subsection{Evaluation}
We first evaluate the prediction accuracy of the behavioral cloning on both classification and regression tasks, by measuring the Root Mean Square Error (RMSE) for the regression task (with the continuous action space) and the Area Under the ROC curve (AUROC) for the classification task (with the discrete action space), respectively (See section \ref{learn} for formulation). We got the reliable performance of behavioral cloning (Table \ref{bc_auc}), which we can use to learn the expert policy in a patient subgroup and estimate the differences in another patient subgroup in a supervised manner. For our future work, we will provide a quantitative distance measure between the counterfactual and original policy trajectory, as well as qualitatively present both of the trajectories. This evaluation will also be reported for the IL (GAIL) and IRL+Offline RL experiments as well.

\begin{table*}[ht!]
	\small
	\centering
	\caption{Performance of the Feed Forward Network (FFW) on Behavioral Cloning of Expert Action.}
	\begin{tabular}{ccc}
 		\toprule
 		& \multicolumn{2}{c}{FFW} \\
 		\cmidrule(r){2-3}
		Outcome label & AUROC & RMSE \\
		\midrule
        Binned Action Classification & 0.83 $\pm$ 0.01 & - \\
        Continuous Action Regression: Fluid & - & 0.68 $\pm$ 0.05 \\
        Coninuous Action Regression: Vasopressor & - & 0.41 $\pm$ 0.06\\
		\bottomrule
	\end{tabular}
\label{bc_auc}
\end{table*}

\section{Future Works and Broader Impact}
In this work, we used several reinforcement learning methods to have the model learn the optimal expert treatment policy on sepsis-3 and community-onset sepsis patient cohort. We aim to identify the difference in action trajectory according to each subpopulation by comparing the output counterfactual policy of the learned models and real expert action trajectory. We would like to achieve this in several subpopulations, such as different gender (Male, Female) groups, ethnic groups (White, Black), patients treated under different guidelines, and patients holding different insurance plans. The differences in care will further be identified and evaluated in pre-ICU triage care (MGB dataset) as well as in ICU (MIMIC-IV, MGB dataset), by which we will be able to contribute to fair treatment planning. This is meaningful and impactful in that we usually don’t have the counterfactual, 'what if' situation such as: what if this patient is under a different insurance plan or what if this patient has been treated under different guidelines. We would like to achieve this goal by leveraging several frameworks such as behavioral cloning, imitation learning, and the combination of inverse reinforcement learning and offline reinforcement learning.

Given distinct sub-populations with the deviation of treatment policies from counterfactual policies, we can analyze what drives this deviation from the other direction of action. Our result can also provide qualitative insights to the questions related to causality or fairness issues, such as: what feature \emph{causes} the treatment change? or is the treatment change due in part to the ethnicity of a patient or any other clinical features? Answering these questions would help us better utilize recent advances in reinforcement learning and deep learning models to understand real-world datasets.

\textbf{Limitations and Concerns} We are using the state-transition model used for the rollouts for IL which has been trained based on patient treatment history from observation data. Thus, our state-transition model does not guarantee its robustness to the patient state and expert treatment policy with different distribution from the input data. Furthermore, as proven by Swamy et al., \cite{swamy2022sequence} this history-dependent policies generated by IL might work well with off-policy IL such as behavioral cloning but might work relatively poorly on on-policy IL approach.

\section{Acknowledgements}
We would like to acknowledge and thank our sponsors, who support our research with financial and inkind contributions: S. Nayak was sponsored in part by the United States AFRL and the United States Air Force Artificial Intelligence Accelerator and was accomplished under Cooperative Agreement Number FA8750-19-2-1000. We would like to thank members of the HealthyML Lab for their invaluable feedback.

\bibliographystyle{plain}
\bibliography{reference}

\clearpage
\appendix
\section{Visualization of action in patient subgroups}
\begin{figure*}[t!]
\centering
\subfigure[Vasopressor Treatment for all patients]{\includegraphics[width=0.49\textwidth]{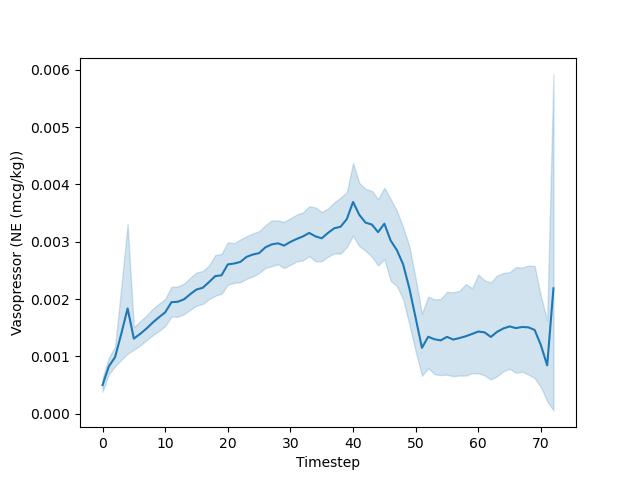}}
\subfigure[Fluids Treatment for all patients]{\includegraphics[width=0.49\textwidth]{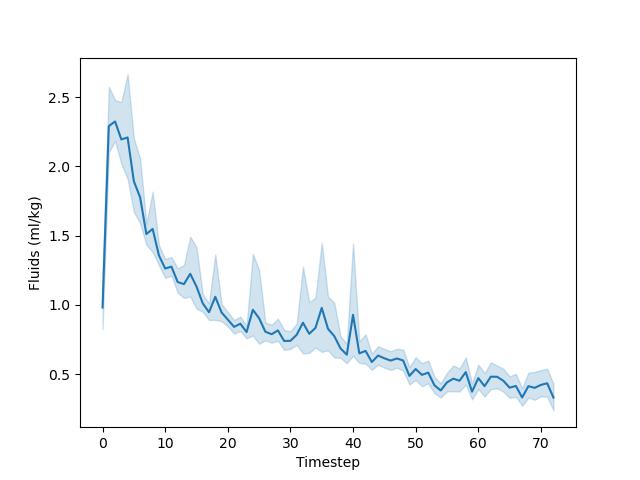}}\\
\subfigure[Vasopressor Treatment in gender subgroups]{\includegraphics[width=0.49\textwidth]{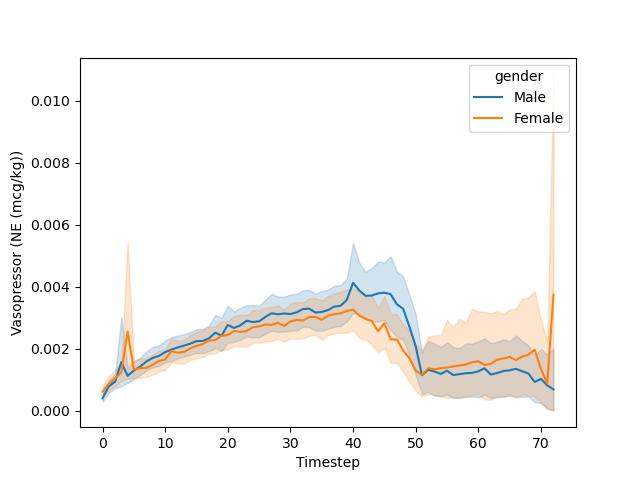}}
\subfigure[Fluids Treatment in gender subgroups]{\includegraphics[width=0.49\textwidth]{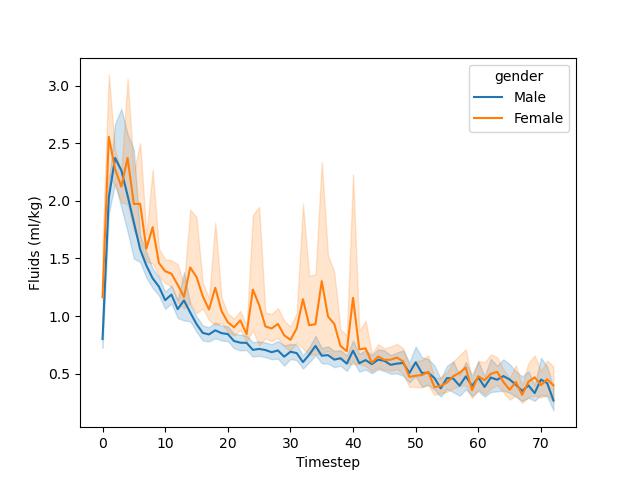}}\\
\subfigure[Vasopressor Treatment in different racial subgroups]{\includegraphics[width=0.49\textwidth]{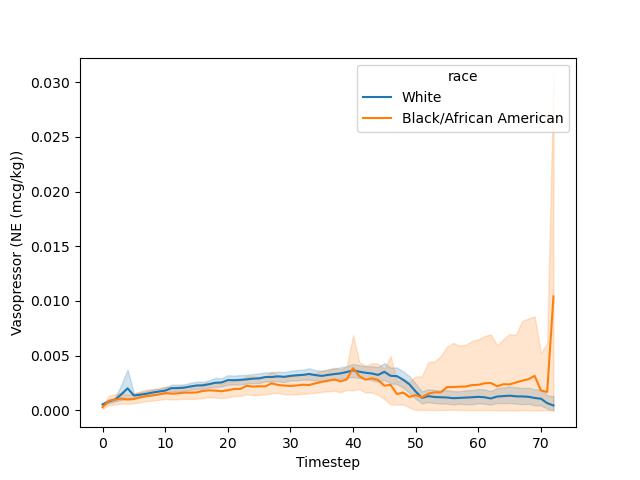}}
\subfigure[Fluids Treatment in different racial subgroups]{\includegraphics[width=0.49\textwidth]{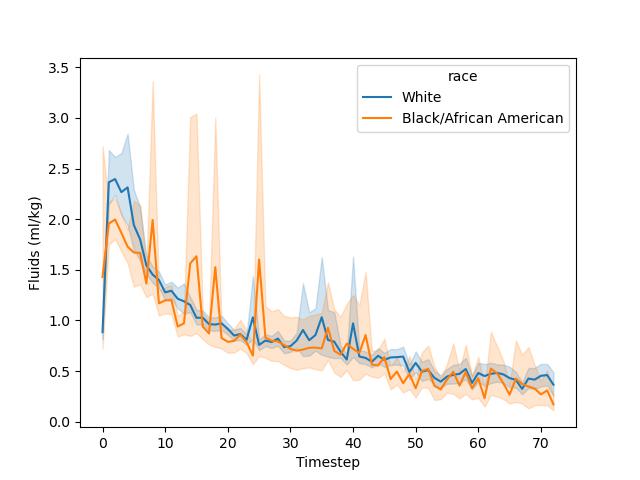}}
\caption{\small \textbf{Action Trajectory of Patients across timestep}}
\label{action_trajectory}
\end{figure*}

We visualize the action trajectory of each patient subgroup to see the inherent differences in care. Figure \ref{action_trajectory} summarizes the vasopressor (Figure \ref{action_trajectory} (a), (c), (e)) and Fluids treatment amount (Figure \ref{action_trajectory} (b), (d), (f)) across timestep, according to different gender (Figure \ref{action_trajectory} (c), (d)) and race (Figure \ref{action_trajectory} (e), (f)) group. Here the vasopressor treatment is the norepinephrine equivalent amount explained in Section \ref{conti_action}, also fluid amount has been calculated considering the different effects of fluids. We could qualitatively observe the difference in care across different gender groups, specifically for vasopressor treatment, where female patients generally tend to get less amount of vasopressor, where the fluid amount was generally similar across different timestep except later in the treatment course. Different racial groups also were treated with different patterns of treatment courses in both vasopressors and fluids. Further clinical investigations and interpretation will be followed for a more detailed explanation of this drift in treatment policy.

\end{document}